# STREAMING OBJECT DETECTION ON FISHEYE CAMERAS FOR AUTOMATIC PARKING

[1]Yixiong, Yan*; [1]Liangzhu, Cheng; [1]Yongxu, Li; [1]Xinjuan, Tuo; [1]Baoqiang, Huang; ; [1]Yakun, Zhu;

[1]Dongfeng Motor Group Co., Ltd. Technology Center, Wuhan, China

ABSTRACT

Fisheye cameras are widely employed in automatic parking, and the video stream object detection (VSOD) of the fisheye camera is a fundamental perception function to ensure the safe operation of vehicles. In past research work, the difference between the output of the deep learning model and the actual situation at the current moment due to the existence of delay of the perception system is generally ignored. But the environment will inevitably change within the delay time which may cause a potential safety hazard. In this paper, we propose a real-time detection framework equipped with a dual-flow perception module (dynamic and static flows) that can predict the future and alleviate the time-lag problem. Meanwhile, we use a new scheme to evaluate latency and accuracy. The standard bbox is unsuitable for the object in fisheye camera images due to the strong radial distortion of the fisheye camera and the primary detection objects of parking perception are vehicles and pedestrians, so we adopt the rotated bbox and propose a new periodic angle loss function to regress the angle of the box, which is the simple and accurate representation method of objects. The instance segmentation ground truth is used to supervise the training. Experiments demonstrate the effectiveness of our approach. Code is released at: https://gitee.com/hiyanyx/fisheye-streaming-perception.

## 1. INTROUDUCE

For automatic parking, the fisheye camera is an essential sensor, and the video stream object detection (VSOD) of the fisheye camera has become a fundamental perception function, providing important information for obstacle avoidance and path planning. In past research work, many excellent object detectors [1] can achieve good performance and it will always take a certain amount of time for the detector to infer an image frame, about a few milliseconds to tens of milliseconds. However, the difference between the output of the deep learning model and the actual situation at the current moment due to the delay of the perception system is generally ignored. But the environment will inevitably change within the delay time which may cause a potential safety hazard. To alleviate the time-lag problem on automatic parking, VSOD should have the ability to future forecasting.

To facilitate research on VSOD, the stream average precision (sAP) is proposed which is a novel metric to integrate delay and accuracy into a single metric for evaluating video stream object detection and is referred to as "stream accuracy"[3]. Literature [3] shows that the performance of many excellent object detectors decreases significantly based on sAP evaluation because the delay of the perception system causes the current frame result of the detector to always match the next frame. Those detectors use the heuristic method can restore certain sAP performance [8, 4], which can realize knowing the history state to predict the result of the next moment. **In this paper**, rather than equipped with the heuristic method, we propose a real-time detection framework equipped with a dual-flow perception (DFP) module (dynamic and static flows) which can predict the future and alleviate the time-lag problem.

The fisheye camera is an essential sensor and the inherent geometric transformation of the fisheye camera will lead to a massive change in space distortion. The wide field of view of the fisheye image is accompanied by a strong side effect of radial distortion. Objects at different angles to the optical axis look very different. Therefore, the detector should regress an appropriate object bbox which is of great research significance. The representation methods include (1) Standard bbox. (2) Polygon bbox. (3) Rotated bbox. FisheyeDet is represented by 4-sided polygons [10]. FisheyeYOLO[8] explored curved boxes and adaptive stepped polygons. 24 points polygonal bbox is used to represent [9]. **In this paper**, we adopt the rotated bbox which is the simple and accurate representation method of objects, because the primary detection objects of parking perception are vehicles and pedestrians for which rotated bbox is suitable. The regression of the polygonal bbox is complex and difficult to converge. The standard bbox will contain more redundant background information and cannot provide accurate location and inaccurate annotation, on distorted images.

For angle regression, traditional L1/L2 loss is often simply used [16, 24]. Because of angle symmetry problems, the performance is average. **In this paper**, we propose a new periodic angle loss function to regress the angle.

In this paper, the contribution includes the following aspects:

(1) VSOD should regress an appropriate bbox and predict the future, which is of great significance for the auto parking system which uses the fisheye camera as the essential sensor.
(2) DFP module is equipped to realize the ability to predict the future and alleviate the time-lag problem.
(3) Rotated bbox is the simple and accurate representation method of objects in fisheye images and a new periodic angle loss function is used to regress the angle of bbox, which is better than the traditional regression loss function and does not introduce additional complex calculations.

## 2. RELATED WORK

In this chapter, we analyze the previous work on video stream detection and the bbox representation method.

2.1 VIDEO STREAM DETECTION

**Image object detection**. The detection algorithm based on depth learning can be divided into two stages and one stage framework. Many excellent works consider both speed and accuracy [1,13].

**Offline video stream object detection**[11,12]. The purpose is to achieve a stable detection effect through some methods. Optical flow[26] and other schemes are used to aggregate the features of the time context to deal with occlusion, motion blur, and other work scenes. Those ignore the result of the perception system and the actual situation at the current time because of the delay of inference.

**Online video stream object detection**[3]. It considers the difference between the real-world situation and the model inference result caused by delay, and realizes the balance between delay and accuracy. Online video stream perception was expected to know and report the state of the world at any time, even if they did not process the previous frame. The sAP integrates delay and accuracy into a single metric and evaluates the results in a continuous time range according to the current real-world truth value including the effect of the delay. Some detectors use the heuristic method (such as the Kalman filter) can restore certain sAP performance [8, 4], which can realize knowing the history state to predict the result of the next moment. In this paper, we focus on **online video stream object detection**.

2.2 BBOX REPRESENTATION

Due to the inherent geometric transformation of the fisheye camera leads, it is necessary to find more appropriate mathematical schemes to accurately represent the objects in the image. There are two ways: (1) method based on distortion correction; (2) method based on the original image.

2.2.1 Method Based on Distortion Correction

With the undistorted pinhole camera projection, a straight line in the real-world is still a straight line in the image after the pinhole mapping. However, with the fisheye camera projection, such as a camera with an isometric projection lens, a line in the real-world will be transformed and displayed as a curve in the image.

It seems inappropriate to directly use the standard bbox on the fisheye image, but the scheme based on distortion correction comes naturally. After distortion correction, fisheye images similar to pinhole imaging can be generated. The object detection method is similar to the standard pinhole camera scheme. This method usually includes two stages: image distortion and object detection. Image deformation is the key to these methods. A common practice is to use the fourth-order polynomial [5] model or the unified camera model [6] to correct the distortion in the image. Note that, for fisheye cameras, with resampling distortion, especially in the periphery of the image, on the one hand, it is understandable that the introduction of false frequency components harms computer vision [7]; On the other hand, other smaller effects include the reduced field of view due to invalid pixels and non-rectangular images [8].

2.2.2 Method Based on Original Image

**Semantic segmentation and instance segmentation** can help to obtain object contours, but these tasks are computationally complex and usually require a bbox estimation step.

**Standard bbox** is commonly used to represent objects. They are represented by four parameters $(x, y, w, h)$, representing the center, width, and height of the box. The bbox aligns with the pixel grid. This method is effective for pinhole camera images without distortion. But for fisheye distorted images, this method can not accurately locate the obstacle and it captures a lot of redundant background information which is bad for training.

**Rotated bbox** is an extension of the standard bbox with additional parameters θ to capture the rotation angle of the box. Traditionally, regression functions based on L1/L2 distance. It is commonly used in lidar top-looking object detection methods [14]. Previous research has taken a different approach. For examples, RSDet [16] uses the point-based eight-parameter detection method to sort the bbox corners and MultiBin[15] discretizes the target range and divides it into n overlapping bins. **In this paper, the focus is on the improvement of L1/L2 for periodic problems and does not introduce additional complex calculations**, so our method is different from the ideas in literature 15 and 16.

**Rotated ellipse** shares the same parameter type as the rotated bbox. The advantage of a rotated ellipse is that the edge area is small, which is more suitable for some special scenes. Ellipse R-CNN [18] uses ellipses to represent objects rather than boxes.

**Polygons** can represent any shape. It can be divided into three representations. (1) Polygons with uniform angle sampling [19,20]. Based on the polar coordinate system. 24 points polygonal bbox is used to represent [9]. (2) Polygons with a uniform sampling of contour circumference. Based on the Cartesian coordinate system. (3)Polygons with adaptive sampling polygon based on local contour curvature[8]. Based on the Cartesian coordinate system. More vertices are used in areas of higher curvature than lines, and lines can be represented by smaller vertices. The algorithm[21] is to detect the main points of the object. The algorithm[22] to reduce the points to obtain the simplified curve.

## 3. NETWORK STRUCTURE

The model follows the paradigm of a one-stage detector, including a backbone, a feature pyramid network (FPN), a dual-flow perception module, and a detector head. The detector head is a rotated bbox regression. The model structure is shown in **Figure 1**, where arrows represent multiple convolution layers, and the color rectangles represent feature maps, whose size corresponds to the size of $h \times w = 640 \times 640$. The pipeline can be expressed as, seen in the equations:

$$P_{\{k\}}^{\{fpn@t\}} = \text{BackboneFPN}(I^{\{@t\}}) \; \forall k = 1,2,3 \dots \; (1)$$

$$P_{\{k\}}^{\{fpn@t-1\}} = \text{BackboneFPN}\left(I^{\{@t-1\}}\right) \; \forall k = 1,2,3 \ldots (2)$$

$$F_{\{k\}}^{\{DFP\}} = \text{DFP}\left(P_{\{k\}}^{\{fpn@t\}}, P_{\{k\}}^{\{fpn@t-1\}}\right) \; \forall k = 1,2,3 \ldots (3)$$

$$\widehat{T_{\{k\}}} = \text{Head}\left(F_{\{k\}}^{\{DFP\}}\right) \; \forall k = 1,2,3 \ldots (4)$$

Where, BackboneFPN(·) represents the operation of an image through a backbone network and a feature pyramid network. k is the resolution level and the network has a total of three resolutions. @t is the current moment and @t − 1 is the historical moment. $I^{\{@t\}}$ and $I^{\{@t-1\}}$ are the input image frames of the current moment and the historical moment. $P_{\{k\}}^{\{fpn@t\}}$ and $P_{\{k\}}^{\{fpn@t-1\}}$ are the output multi-dimensional feature matrix of the BackboneFPN(·) operation. DFP(·) represents the operation of the dual-flow perception module. $F_{\{k\}}^{\{DFP\}}$ is the output of the DFP(·) operation. $\widehat{T_{\{k\}}}$ represents the prediction information at different resolution levels. And in this article, the predicted value is represented by adding a hat above the letter. For example, $\widehat{T}$ represents the predicted value and T represents the ground truth.

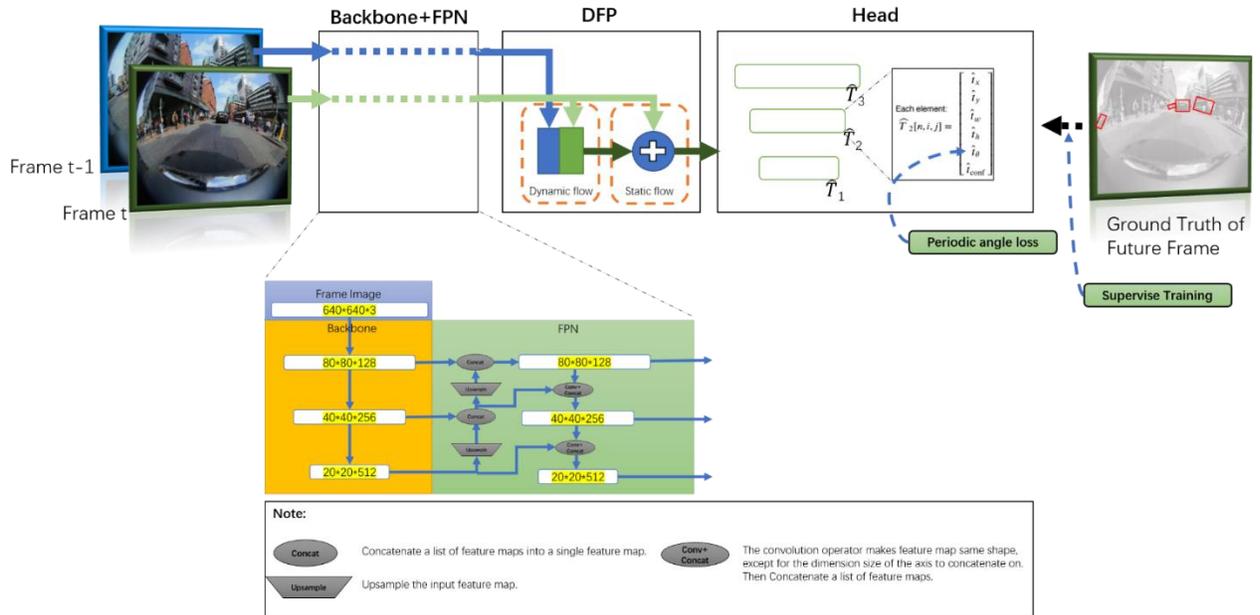

Figure 1. Network architecture. Arrows represent multiple convolution layers and the color rectangles represent feature maps. We chose the CSPDarknet backbone and FPN used in YOLOX-s to extract features of image frames. Use Dual-Flow Perception Module to fuse feature maps. Ground truth of future frames to supervise the training. In the detection head part, the periodic angle loss function is used for the angle prediction of the rotated bbox.

3.1. BACKBONE AND FPN NETWORK

Due to that, we focus on **online video stream object detection** on designing better methods to boost the detector to predict the future but not on reducing the FLOPS of the backbone network. In this paper, the Backbone network and FPN network we selected come from YOLOX, which is the CSPDarknet and FPN used in YOLOX-s[2]. YOLOX achieved good results in CVPR 2021 Streaming Perception Challenge. **To reduce the experimental burden, we did not compare other backbone networks, but focused on the DFP module and a new periodic angle loss function**.

## 3.2 DUAL-FLOW PERCEPTION MODULE

The current image frame $I^{\{@t\}}$ and historical image frame $I^{\{@t-1\}}$ are used as input, and the current feature map $P^{\{fpn@t\}}$ and historical feature map $P^{\{fpn@t-1\}}$ are obtained through the backbone network and FPN network. Considering that these two features contain the key information of the time context and can be used to predict the future information of the next frame, we designed a dual-flow perception module to fuse these two feature maps. The specific operation is divided into two steps.

The first step is that gdynamic(·) makes a dynamic flow fuses to learn motion information. For the fusion of dynamic flows, the two feature maps $P^{\{fpn@t\}}$ and $P^{\{fpn@t-1\}}$ employ a shared weight $1 \times 1$ convolution layer respectively, then use batch norm and SiLU to reduce the number of channels of the two FPN features to half, and then perform the concatenation operation to generate the dynamic feature $F_{\{dynamic\}}$. After the above analysis, we can write the following formula (5).

$$F_{\{dynamic\}} = gdynamic(P^{\{fpn@t\}}, P^{\{fpn@t-1\}}) \ldots (5)$$

The second step is that gstatic(·) makes a static flow fuse to get basic feature information of the current moment. For the fusion of static flow, the dynamic flow $F_{\{dynamic\}}$ and the features $P^{\{fpn@t\}}$ of the current frame are connected by reasonable residuals to obtain the dual-flow feature $F^{\{DFP\}}$. After the above analysis, we can write the following formula (6).

$$F^{\{DFP\}} = gstatic(P^{\{fpn@t\}}, F_{\{dynamic\}}) \ldots (6)$$

Through these two steps, the basic information of the current frame is obtained, the motion trend is learned, and the robustness of the detection is improved.

## 3.4 DETECTION HEAD

For the bbox representation method, we use the rotated bbox.

The detection head is an **anchor-based** scheme that predicts the bbox at 3 different resolution levels and predicts 3 boxes at each grid cell. The output of the detection head is $\widehat{T_{\{k\}}}_{\{k=1\}}^{\{3\}}$, which represents the prediction information and k is the resolution level, so the output tensor of the detection head at each grid cell is $[3 * (4 + 1 + 1)]$ for the 4 bbox offsets, 1 angle prediction, and 1 confidence prediction. In the location coordinates relative to each grid cell, the $n^{\{th\}}$ predicted rotated bbox $\widehat{T_{k[n,i,j]}} = (\widehat{\{tx\}}, \widehat{\{ty\}}, \widehat{\{tw\}}, \widehat{\{th\}}, \{tangle\}, \{tconf\}) \in R^6, n \in [1,2,3]$ of the cell (i, j) at the resolution level k. $\widehat{\{tx\}}$, $\widehat{\{ty\}}$ the value range is [0,1].

In the image coordinates, the upper left point of the image is the origin. $b = (\widehat{\{bx\}}, \widehat{\{by\}}, \widehat{\{bw\}}, \widehat{\{bh\}}, \{bangle\}, \{bconf\}) \in R^6$ to denote a prediction the rotated bbox. $\widehat{\{bx\}}, \widehat{\{by\}}$ is the **center point** of the bbox, $\widehat{\{bw\}}, \widehat{\{bh\}}$ is the width and height of the bbox, $\{\widehat{bangle}\}$ is the angle of the bbox and the definition of $\{\widehat{bangle}\}$ is detailed in "3.6 Chapter Periodic angle loss function".

we can write the following formulas (7-12).

$$\widehat{\{bx\}} = \{s_k\}\left(i + Sig(\widehat{\{tx\}})\right) \ldots (7)$$

$$\widehat{\{by\}} = \{s_k\}(j + Sig(\widehat{\{ty\}})) \ldots (8)$$

$$\widehat{\{bw\}} = \{w_{\{k,n\}}^{\{anchor\}}\}e^{\{tw\}} \ldots (9)$$

$$\widehat{\{bh\}} = \left\{h_{\{k,n\}}^{\{anchor\}}\right\}e^{\{th\}} \quad \ldots (10)$$

$$\widehat{\{bangle\}} = \alpha \text{Sig}(\widehat{\{tangle\}}) - \beta \quad \ldots (11)$$

$$\widehat{\{bconf\}} = \text{Sig}(\widehat{\{tconf\}}) \quad \ldots (12)$$

Here, Sig(·) is a logistic activation function, which is used to constrain the $\widehat{\{tx\}}$, $\widehat{\{ty\}}$ in the classical anchor-based scheme[32]. $\{s_k\}$ as the stride at resolution level k. i and j are the positions of the grid cell. $\left\{w_{\{k,n\}}^{\{anchor\}}\right\}$ and $\left\{h_{\{k,n\}}^{\{anchor\}}\right\}$ are the size of the anchor box at resolution level k. $\widehat{\{bangle\}}$ is limited to the range of $[-\beta, \alpha - \beta]$ whose $\alpha$ and $\beta$ are super parameters.

3.5 LOSS FUNCTION

To reduce the burden, the experiment only considers the category of vehicle, so there is no loss function designed for the category. The loss function refers to RAPiD[17]. we can write the following loss formulas (13-17).

$$L_{\{txty\}} = \sum_{\{\widehat{\{t\}} \in T^{\{pos\}}\}} \left\{\text{BCE}\left(\text{Sig}(\widehat{\{tx\}}), tx\right) + \text{BCE}\left(\text{Sig}(\widehat{\{ty\}}), ty\right)\right\} \quad \ldots (13)$$

$$L_{\{twth\}} = \sum_{\{\widehat{\{t\}} \in T^{\{pos\}}\}} \left\{(\{tw\} - \widehat{\{tw\}})^2 + (\{th\} - \{\widehat{\{th\}}\})^2\right\} \quad \ldots (14)$$

$$L_{\{bangle\}} = \sum_{\{\widehat{\{t\}} \in T^{\{pos\}}\}} \{l_{\{angle\}}\}(\{bangle\}, \widehat{\{bangle\}}) \quad \ldots (15)$$

$$L_{\{bconf\}} = \sum_{\{\widehat{\{t\}} \in T^{\{pos\}}\}} \left\{\text{BCE}(1, \widehat{\{bconf\}})\right\} + \sum_{\{\widehat{\{t\}} \in T^{\{neg\}}\}} \{0, \widehat{\{bconf\}}\} \quad \ldots (16)$$

$$L_{\{total\}} = L_{\{txty\}} + L_{\{twth\}} + L_{\{bangle\}} + L_{\{bconf\}} \quad \ldots (17)$$

Where $L_{\{total\}}$ is the total loss of bbox regression minimization. $T^{\{pos\}}$ and $T^{\{neg\}}$ are positive and negative samples and calculation method is easy to understand.

3.6 PERIODIC ANGLE LOSS FUNCTION

Traditionally, L1/L2 distance are used for angle prediction. Look at formulas (18), $L_{\{bangle-normal\}}$ loss function L1 norm. **Figure 2** is $L_{\{bangle-normal\}}$ L1 norm form.

$$L_{\{bangle-normal\}} = \sum_{\{\widehat{\{t\}} \in T^{\{pos\}}\}} \text{abs}(\{bangle\} - \widehat{\{bangle\}}) \quad \ldots (18)$$

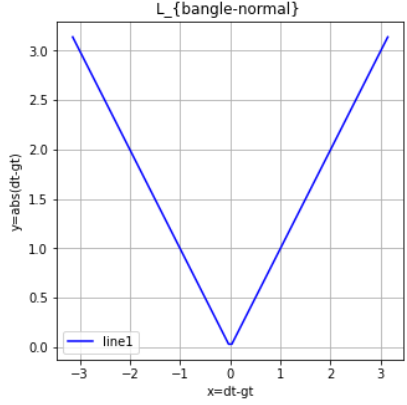
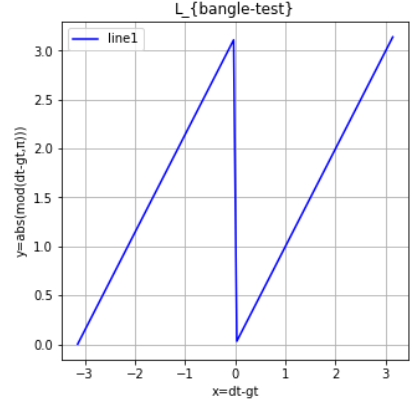

Figure 2. $L_{\{bangle-normal\}}$ L1 norm form. "dt" is $\{\widehat{bangle}\}$. "gt" is $\{bangle\}$. The abscissa is x=dt-gt and the ordinate is $y = abs(dt - gt)$. The curve visualizes the loss function of $L_{\{bangle-normal\}}$.

Figure 3. $L_{\{bangle-test\}}$ L1 norm form. "dt" is $\{\widehat{bangle}\}$. "gt" is $\{bangle\}$. The abscissa is x=dt-gt and the ordinate is $y = abs(mod(dt - gt, \pi))$. The curve visualizes the loss function of $L_{\{bangle-test\}}$.

### 3.6.1 The First Problem

$\{bx\}, \{by\}, \{bw\}, \{bh\}, \{bangle\}$ and $\{bx\}, \{by\}, \{bw\}, \{bh\}, \{bangle + \pi\}$ can represent the same rotated bbox, so the periodic angle loss function must meet this requirement, formula (19):

$$\text{Lossfunction}(\{\widehat{bangle}\}, \{bangle\}) = \text{Lossfunction}(\{\widehat{bangle}\}, \{bangle + \pi\}) \ldots (19)$$

For the first case, we thought of using $\{\{\widehat{bangle}\} - bangle\}$ and $\pi$ to do the remainder operation to ensure that the loss function is a periodic function of $\pi$. Look at formulas (20-22), $L_{\{bangle-test\}}$ loss function L1 norm and L2 norm form. **Figure 3** is $L_{\{bangle-test\}}$ L1 norm form.

$$\delta\{angle\_test\} = \{\widehat{bangle}\} - bangle \ldots (20)$$

L1 norm form:

$$L_{\{bangle-test\}} = \sum_{\{\{\widehat{t}\} \in T^{\{pos\}}\}} \{abs(mod(\delta\{angle\_test\}, \pi))\} \ldots (21)$$

Where mod $(\cdot)$ represents modular operation (remainder).

The loss function (21) and (22) are unsuitable for the situation of $\{\{\widehat{bangle}\} - bangle\} \approx -0$, which will produce a large gradient. Therefore, we adjust the periodic function to ensure that the value of the loss function is relatively small in this case. Look at formulas (23-24), $L_{\{bangle\}}$ loss function L1 norm and L2 norm form. Figure 4 is $L_{\{bangle\}}$ L1 norm form.

$$\delta\{angle\} = \{\widehat{bangle}\} - bangle - \pi/2 \ldots (23)$$

L1 norm form:

$$L_{\{bangle\}} = \sum_{\{\{\widehat{t}\} \in T^{\{pos\}}\}} \{abs(mod(\delta\{angle\}, \pi) - \pi/2)\} \ldots (24)$$

Where mod $(\cdot)$ represents modular operation (remainder).

Note that, when $\delta\{angle\} = k * \pi, k \in Z$ because $L_{\{bangle\}}$ is nondifferentiable, these situations are ignored in the backpropagation process.

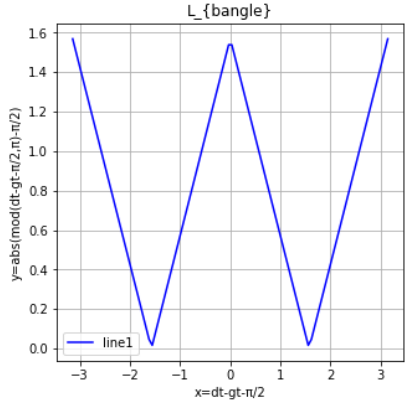
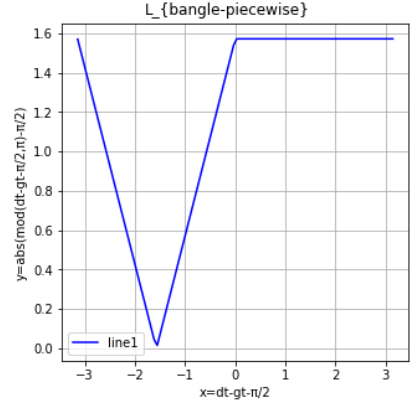

Figure 4. $L_{\{bangle\}}$ L1 norm form. "dt" is $\widehat{\{bangle\}}$. "gt" is $\{bangle\}$. The abscissa is x=dt-gt-π/2 and the ordinate is $y = abs(mod(dt - gt - \pi/2, \pi) - \pi/2)$. The curve visualizes the loss function of $L_{\{bangle\}}$.

Figure 6. $L_{\{bangle-piecewise\}}$ L1 norm form. "dt" is $\widehat{\{bangle\}}$. "gt" is $\{bangle\}$. The abscissa is x=dt-gt-π/2 and the ordinate is a piecewise function. The curve visualizes the loss function of $L_{\{bangle-piecewise\}}$.

To speed up the convergence of the model, we also propose a piecewise functional form of **the piecewise periodic loss function $L_{\{bangle-piecewise\}}$** based on **the periodic loss function $L_{\{bangle\}}$**. The piecewise loss function is that when **$0 < \delta\{angle\} < \pi/2$**, we maintain a relatively large descending gradient, see equation 26. Figure 6 is $L_{\{bangle-piecewise\}}$ L1 norm form.

L1 norm form:

$$L_{\{bangle-piecewise\}} = \sum_{\{\widehat{\{t\}} \in T^{\{pos\}}, \delta\{angle\}<0\}} \left\{ abs\left(mod(\delta\{angle\}), \pi\right) - \frac{\pi}{2}\right\} + \sum_{\{\widehat{\{t\}} \in T^{\{pos\}}, \delta\{angle\}>0\}} \left\{abs\left(mod(0, \pi) - \frac{\pi}{2}\right)\right\} \dots (26)$$

3.6.2 the Second Problem

$\{bx\}, \{by\}, \{bw\}, \{bh\}, \{bangle\}$ and $\{bx\}, \{by\}, \{bh\}, \{bw\}, \{bangle + \pi/2\}$ can represent the same rotated bbox.

In order to ensure that the training of the network model can converge, we need to ensure that the label information of the rotated bbox cannot be ambiguous. Therefore, we set the rule that the shorter side of the rotated bbox is the width and the longer side is the height, this is, **$\{bw\} < \{bh\}$.** In this case, the angle range $[-\pi/2, \pi/2)$ of the rotated bbox can unambiguously represent the whole situation. So we modify the annotation of the dataset, force the bbox to bw < bh, and make the angle range to $[-\pi/2, \pi/2)$. As shown in Figure 5.

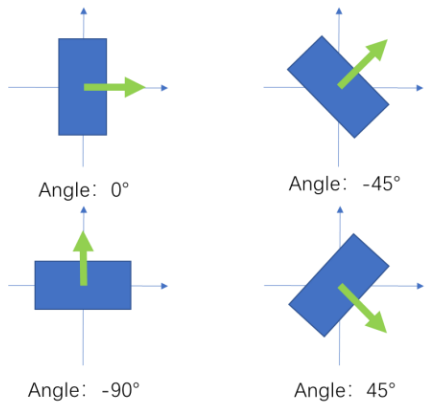 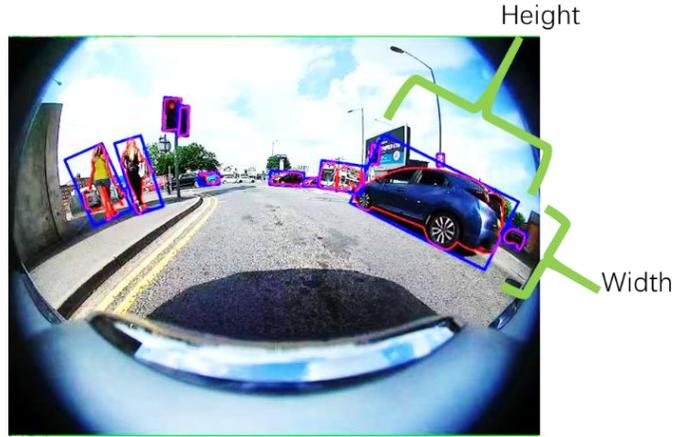

Figure 5. The angle range of the rotation bbox is $[-\pi/2, \pi/2)$, which can describe the whole situation of any rotation of the bbox.

Figure 7. The red outline points are the original labeling information of WoodScape and the blue bbox is the bbox we generate. The shorter side of the rotated bbox is the width and the longer side is the height, this is, **{bw} < {bh}.**

3.6.3 the Super Parameters α and β

In Formula 11, the range of $\widehat{\{bangle\}}$ in angle prediction is $[-\beta, \alpha - \beta]$.

Because we modify the annotation of the dataset, force the bbox to bw < bh, and make the angle range to $[-\pi/2, \pi/2)$. The range $[-\pi/2, \pi/2)$ of {bangle} can represent all cases of rotated bboxes. So we naturally think of the following: α= π， β= π/2.

But in Figure 4, **when $0 < \delta\{angle\} < \pi/2$, the first derivative of $L_{\{bangle\}}$ is negative, which causes the result after backpropagation to be more deviated from the ground truth**. For dealing with this situation, the network should learn to estimate $\widehat{\{bangle\}}$ as {bangle ± π}, ensuring that the range of δ{angle} is between $(-\pi, -\pi/2)$, which is a reasonable backpropagation situation. Therefore, we need to expand the detection range of the detection head $\widehat{\{bangle\}}$ to $[-\pi, \pi]$, and set super parameters α = 2π, β = π.

So far, we propose a reasonable loss function, **the periodic loss function $L_{\{bangle\}}$ and the piecewise periodic loss function $L_{\{bangle-piecewise\}}$**, and we modify the annotation of the dataset, force the bbox to bw < bh, and make the angle range to $[-\pi/2, \pi/2)$. Those ideas can deal with the periodicity problems of the rotated boundary box and speed up the convergence of the model.

3.7 DATASET

3.7.1 Generate Annotation using Instance Segmentation

We have selected two Datasets, one is WoodScape[23], and the other is Argoverse-HD[25]. To reduce the burden, the experiment only considers the category of vehicle.

WoodScape dataset is a fisheye dataset, which is in RGB format, with a resolution of 1MPx and a horizontal FOV of 190 °. Using the Front/ Rear / Mirror Left/ Mirror Right cameras and the corresponding annotation, there are a total of 7,000 pictures in the training dataset and 1,200 pictures in the verification dataset.

Argoverse-HD is a video stream object dataset, which can be used to evaluate the performance of high frame rate object detection and calculate the sAP performance indicators. Use only the central RGB camera and the corresponding annotation. The training dataset has a total of 65 pieces of data, and each piece of data has 900 pictures. The verification collection has a total of 24 pieces of data, and each piece of data has 900 pictures. **The assigned training and verification annotation information are available on our gitee warehouse**.

3.7.2 Preparation of Dataset

WoodScape and Argoverse-HD have no rotated bbox label, we wrote some scripts to generate the rotated bbox. In Figure 7, the red outline points are the original labeling information of WoodScape and the blue bbox is the bbox we generated.

We generate a blue bbox based on the red outline points. The rotated bbox can be divided into two types that are **Minimum Area Bounding Rectangle** and **Minimum Perimeter Bounding Rectangle**. Generally, there is little difference between the two. We choose the type of Minimum Area Bounding Rectangle. The general methods for solving the Minimum Area Bounding Rectangle are that the convex hull of the object image is solved by the Graham scanning method and the method of rotation or projection is to obtain the rectangle. As described in section 3.6.2 the Second Problem, we set the rule that the shorter side of the rotated bbox is the width and the longer side is the height, this is, **{bw} < {bh}.** In this case, the angle range $[-\pi/2, \pi/2]$ of the rotated bbox can unambiguously represent the whole situation. So we modify the annotation of the dataset, force the bbox to bw < bh, and make the angle range to $[-\pi/2, \pi/2)$.

In order to make model training and result testing more convenient, we constructed a triplet dataset. Because ablation experiments are required, we constructed two types of triplet datasets. The constituent unit of the first type triplet is (Ft, Ft-1, Gt), which contain the picture frame of the current time t, the picture frame of the time t-1, and the ground truth of the current time t. The constituent unit of the second type triplet is (Ft, Ft-1, Gt+1), which contains the picture frame of the current time t, the picture frame of the time t-1, and the ground truth of the future time t+1. Using a triplet dataset can easily shuffle the dataset and execute a more flexible training strategy.

## 4. EXPERIMENTS

4. 1 PERFORMANCE METRICS

We use a bipolar matching method based on a greedy algorithm to calculate the IoU between the detection result of the model and the ground truth. The algorithm is adapted from MS COCO [27]。

4.1.1 Offline Detection Performance based on (Ft, Ft-1, Gt) Triplet Dataset

Offline detection performance. The triplet constituent unit of the trained dataset is (Ft, Ft-1, Gt) and verification is also (Ft, Ft-1, Gt).

Based on the ground truth of time $t$, **{bconf} > 0.01** and IoU_threshold $\in$ {0.5,0.55,0.6,0.65,0.7,0.75,0.8,0.85,0.9,0.95}, calculate the values of TP, FP, and FN, Precision, Recall, sF1, sAP0.5, and sAP0.75 , as well as sAP-mean values. Precision, Recall, and F1 values calculation see formula (27-29).

$$\text{Precision} = \frac{\text{TP}}{(\text{TP} + \text{FP})} \ldots (27)$$

$$\text{Recall} = \frac{\text{TP}}{(\text{TP} + \text{FN})} \ldots (28)$$

$$F1 = \frac{2 * (\text{Precision} * \text{Recall})}{(\text{Precision} + \text{Recall})} \ldots (29)$$

AP: Average Precision is the area of the curve formed by Precision and Recall. AP0.5 and AP0.75 are the AP values under IoU_threshold =0.5 and IoU_threshold =0.75, two typical IoU_thresholds. The AP-mean value is the average value of the AP from IoU_threshold = 0.5 to 0.95.

4.1.2 Online Detection Performance based on (Ft, Ft-1, Gt+1) Triplet Dataset

Online detection performance. The triplet constituent unit of the trained dataset is (Ft, Ft-1, Gt+1) and verification is also (Ft, Ft-1, Gt+1). sAP-mean values. sPrecision, sRecall, and sF1 values calculation see formula (30-32).

$$sPrecision = \frac{sTP}{(sTP + sFP)} \ldots (30)$$

$$sRecall = \frac{sTP}{(sTP + sFN)} \ldots (31)$$

$$sF1 = \frac{2 * (sPrecision * sRecall)}{(sPrecision + sRecall)} \ldots (32)$$

4. 2 MODELS DETAILS

4.2.1 Models Details

We designed a total of **4** sets of models(Model 0-BASE,Model 1 ,Model 2 ,Model3).

For benchmark model, we chose the FisheyeYOLO[8] model trained on the WoodScape dataset. The backbone is ResNet18 with the FPN, and the detection head outputs a standard bbox. We have no additional fine-tuning training. Although this benchmark comparison has some flaws, because the backbone networks used are inconsistent, it still shows the mainstream detection level of the current standard bbox on the fisheye picture. By comparing with the benchmark model, it can also be shown that the solution we proposed has a significant performance improvement.

Table 1. By controlling variables, the designed models are as follows.

| Model Name | CSPDarknet +FPN | DFP Model | $L_{\{bangle-normal\}}$ | $L_{\{bangle\}}$ or $L_{\{bangle-piecewise\}}$ |
|---|---|---|---|---|
| benchmark | × | × | √ | |
| Model 0-BASE | √ | × | √ | |
| Model 1 | √ | × | | √ |
| Model 2 | √ | × | | √ |
| Model 3 | √ | √ | | √ |

4.2.2 Implementation Details

The experimental pre-training weights were obtained from COCO pre-trained model by 15 epochs. For the WoodScape dataset, the models are trained 10,000 iterations (one iteration contains 128 images). The anchor configuration data is used as anchors1 = [24, 45], [28, 24], [50, 77]; anchors2 = [52, 39], [92, 145], [101, 69]; anchors3 = [52, 39], [92, 145], [101, 69]. For the Argoverse-HD dataset, the models are trained 30,000 iterations (one iteration contains 128 images). The anchor configuration data is used as anchors1 = [18, 33], [28, 61], [48, 68]; anchors2 = [45, 101], [63, 113], [81, 134]; anchors3 = [91, 144], [137, 178], [194, 250]. Use Stochastic Gradient Descent (SGD) for training. The learning rate is 0.001, batch size is 24, momentum is 0.9, and weight decay is 0.0005. Rotation, flipping, resizing, and color enhancement are not applicable. For inference, we keep the input size at 640×640, and run on GTX 3090ti GPU.

4.2.3 Benchmark Model Details

4. 3 OFFLINE DETECTION PERFORMANCE TEST FOR AP

The purpose of this experiment is to test the performance improvement of **the periodic angle loss function**. Based on (Ft, Ft-1, Gt) Triplet dataset of WoodScape. The truth angle of this dataset is the diversity and using $L_{bangle-piecewise}$ can converge faster. We compared model 0-BASE, model 1 and benchmark model, the results are in Table 2, and the PR diagrams are Figure 8-10.

Table 2. Performance in the WoodScape dataset for Offline Detection Performance Test. Analysis of the performance of the *periodic angle loss function* $L_{bangle-piecewise}$, based on the weight file after the 10,000th iteration.

| | IoU = 0.5 | | | | | | | IoU = 0.75 | | | | | | | IoU = 0.5:0.95 |
|---|---|---|---|---|---|---|---|---|---|---|---|---|---|---|---|
| | TP | FP | FN | Precision | Recall | F1 | AP0.5 | TP | FP | FN | Precision | Recall | F1 | AP0.75 | AP-mean |
| benchmark model | 2152 | 3161 | 3311 | 0.40504 | 0.39392 | 0.39940 | 0.295 | 312 | 5001 | 5151 | 0.05872 | 0.05711 | 0.05790 | 0.011 | 0.080 |
| Model 0-BASE | 2489 | 30993 | 2974 | 0.07434 | 0.45561 | 0.12782 | 0.129 | 214 | 33268 | 5249 | 0.00639 | 0.03917 | 0.01100 | 0.005 | 0.031 |
| Model 1 | 2767 | 3047 | 2696 | 0.47592 | 0.50649 | 0.49073 | **0.379** | 334 | 5480 | 5129 | 0.05745 | 0.06114 | 0.05924 | **0.016** | 0.097 |

In the case of IoU-0.5, compared with the benchmark model, the number of TP in Mode 2 increases by **10.05%**; the number of FP and FN are slightly less than Benchmark, Precision increases by **0.07088**, Recall increases by **0.11257**; F1 increases by **0.09133**; AP0.5 increases by **0.084**. Compared with Model 0-BASE, the number of TP in Mode 2 increases by **28.58%**; the number of FP is significantly less than that of Model 0-BASE, and the number of FN is slightly less than that of Mode 0-BASE. Precision increases by **0.40158**, Recall increases by **0.05088**; F1 increases by **0.36291**; AP0.5 increases by **0.25**. The performance of Model 1 is significantly better than that of the benchmark model and Model 0-BASE. Comparing the results of the benchmark model and Model 0-BASE, in the fisheye picture, using $L_{bangle-normal}$ reduces the target detection performance. In the case of IoU-0.75, compared with the benchmark model, the number of TP in Mode 2 increases by **7%**; Precision is slightly reduced by **0.00127**, Recall increases by **0.00403**; F1 increases by **0.00134**; AP0.5 increased by **0.005**. Compared with Model 0-BASE, Mode 2 all performance is leading. Compared with the benchmark model and Model 0-BASE, Model2's AP-mean has increased by **0.017** and **0.066**, respectively.

The conclusion is that the use of $L_{bangle-piecewise}$ can significantly improve the target detection ability in fisheye pictures. Compared with $L_{bangle-normal}$ loss function, AP-mean is significantly improved. The rotated bbox is better than the standard bbox in AP performance. The Periodic angle loss function we propose is better than the traditional regression loss function and does not introduce additional complex calculations.

01/11/2023

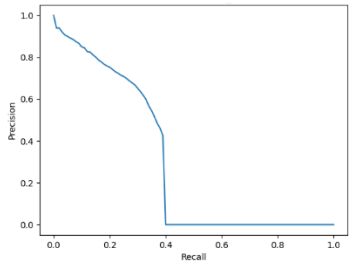
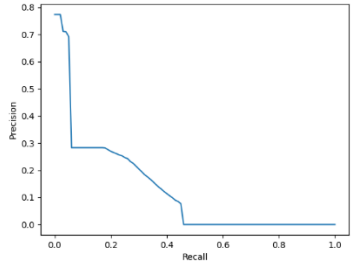
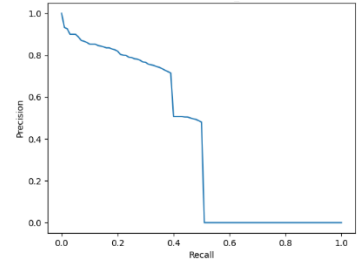

Figure 8. The P-R diagram of benchmark model based on the verification dataset of WoodScape at IoU=0.5, AP0.5 is 0.295, based on the network model and weight file provided by FisheyeYOLO[8].

Figure 9. The P-R diagram of Model 0-Base based on the verification dataset of WoodScape at IoU=0.5, AP0.5 is 0.129, based on the weight file after the 10,000th iteration.

Figure 10. The P-R diagram of Model 1 based on the verification dataset of WoodScape at IoU=0.5, AP0.5 is 0.379, based on the weight file after the 10,000th iteration

4. 4 ONLINE DETECTION PERFORMANCE TEST FOR sAP

The purpose is to test the performance improvement of the DFP module for sAP, based on Argoverse-HD dataset, based on (Ft, Ft-1, Gt+1) Triplet Dataset. Argoverse-HD dataset is a video stream object dataset and can be used to evaluate the performance of high frame rate object detection. We compared model 2 and model 3, using the $L_{\{bangle\}}$ loss function, the results are in Table 3, and the PR pictures are Figure 11-12.

Table 3. Performance in the Argoverse-HD dataset for Online Detection Performance Test. Analysis of the performance of the DFP module, based on the weight file after the 30,000th iteration.

| | IoU = 0.5 | | | | | | | IoU = 0.75 | | | | | | | IoU = 0.5:0.95 |
|---|---|---|---|---|---|---|---|---|---|---|---|---|---|---|---|
| | sTP | sFP | sFN | sPrecision | sRecall | sF1 | sAP0.5 | sTP | sFP | sFN | sPrecision | sRecall | sF1 | sAP0.75 | sAP-mean |
| Model 2 | 55100 | 181166 | 81224 | 0.23321 | 0.40418 | 0.29577 | 0.227 | 18896 | 217370 | 117428 | 0.07998 | 0.13861 | 0.10143 | 0.071 | 0.095 |
| Model 3 | 55280 | 53699 | 81044 | 0.50725 | 0.40550 | 0.45071 | **0.240** | 28615 | 80364 | 107709 | 0.26257 | 0.20990 | 0.23330 | **0.085** | 0.107 |

In the case of IoU-0.5, the number of sTP of Model 2 is the same as the number of sTP of Model 3; the number of sFN of Model 2 is the same as the number of sFN of Model 3; therefore, the sRecall of the two is similar, with only a performance difference of **0.00132**. However, the sFP of Model 2 is larger than the sFP of Model 3, so the sPrecision of Model 3 is **0.27040** higher than that of Model 2, sF1 is **0.15494** higher, and sAP is **0.013** higher.

In the case of IoU-0.75, the number of sTP of Model 2 is about **66%** of the number of sTP of Model 3, and the number of sFN of Model 2 is comparable to the number of sFN of Model3; therefore, the sRecall of Model 3 is **0.071289** higher than that of Model 2. The sFP of Model 2 is larger than the sFP of Model 3, so the sPrecision of Model 3 is **0.18259** higher than that of Model 2, sF1 is **0.13187** higher, and sAP is **0.014** higher.

sAP-mean increased from **0.095** to **0.107**, an increase of **0.012**.

It can be seen that the DFP module has significantly improved sAP performance, and can correctly capture motion trends and make reasonable predictions for the future. **The weight file of the trained model can be obtained in our gitee warehouse**.

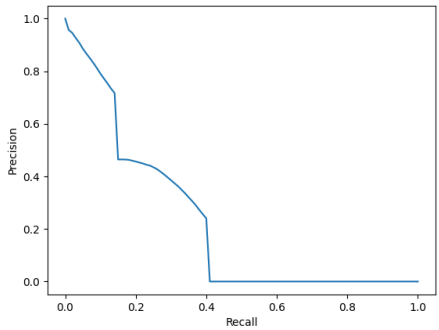
Figure 11. The P-R diagram of Model 2 based on the verification dataset of Argoverse-HD at IoU=0.5, sAP0.5 is 0.227, based on the weight file after the 30,000th iteration

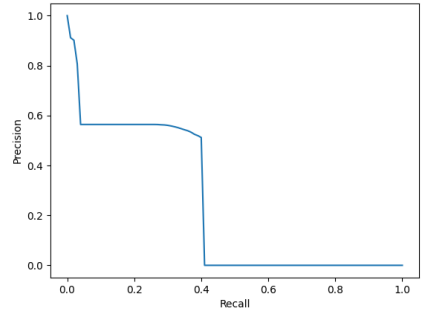
Figure 12. The P-R diagram of Model 3 based on the verification dataset of Argoverse-HD at IoU=0.5, sAP0.5 is 0.240, based on the weight file after the 30,000th iteration.

## 5. CONCLUSION

In this paper, we propose a video stream object detection scheme for the fisheye camera, which adopts a **dual-flow perception module** and a rotated bbox representation method of the object for the fisheye camera. The **periodic angle loss** function is used for the angle prediction of the rotated bbox, which is better than the traditional regression loss function and does not introduce additional complex calculations. We added detectors of the DFP module to achieve a better performance improvement. We are willing to share our experience and code to promote research on fisheye cameras stream perception on.